\documentclass{article}

\usepackage{PRIMEarxiv}

\usepackage[utf8]{inputenc} 
\usepackage[T1]{fontenc}    
\usepackage{hyperref}       
\usepackage{url}            
\usepackage{booktabs}       
\usepackage{amsfonts}       
\usepackage{nicefrac}       
\usepackage{microtype}      
\usepackage{lipsum}
\usepackage{fancyhdr}       
\usepackage{graphicx}       
\usepackage{amssymb}
\usepackage{booktabs}
\usepackage{xcolor}
\usepackage{amsmath}
\usepackage{colortbl}

\newcommand{\etal}{\textit{et al}.}

\title{Laughing Matters: Introducing Laughing-Face Generation using Diffusion Models
}

\author{
  Antoni Bigata Casademunt \\
  Imperial College London \\
  \texttt{ab4522@ic.ac.uk} \\
  \And
  Rodrigo Mira \\
  Imperial College London \\
  \texttt{rs2517@ic.ac.uk} \\
  \AND
  Nikita Drobyshev \\
  Imperial College London \\
  \texttt{nikita.drobyshev23@gmail.com} \\
  \And
  Konstantinos Vougioukas \\
  Imperial College London \\
  \texttt{k.vougioukas@ic.ac.uk} \\
  \And
  Stavros Petridis \\
  Imperial College London \\
  \texttt{stavros.petridis04@ic.ac.uk} \\
  \And
  Maja Pantic \\
  Imperial College London \\
  \texttt{m.pantic@ic.ac.uk} \\
}

\begin{document}
\maketitle

\begin{abstract}
Speech-driven animation has gained significant traction in recent years, with current methods achieving near-photorealistic results. However, the field remains underexplored regarding non-verbal communication despite evidence demonstrating its importance in human interaction. In particular, generating laughter sequences presents a unique challenge due to the intricacy and nuances of this behaviour. This paper aims to bridge this gap by proposing a novel model capable of generating realistic laughter sequences, given a still portrait and an audio clip containing laughter. We highlight the failure cases of traditional facial animation methods and leverage recent advances in diffusion models to produce convincing laughter videos. We train our model on a diverse set of laughter datasets and introduce an evaluation metric specifically designed for laughter. When compared with previous speech-driven approaches, our model achieves state-of-the-art performance across all metrics, even when these are re-trained for laughter generation. Our code and project are publicly available \footnote{\url{https://sites.google.com/view/laughing-matters}}.
\end{abstract}

\begin{figure}[ht]
\centering
\includegraphics[width=0.9\linewidth]{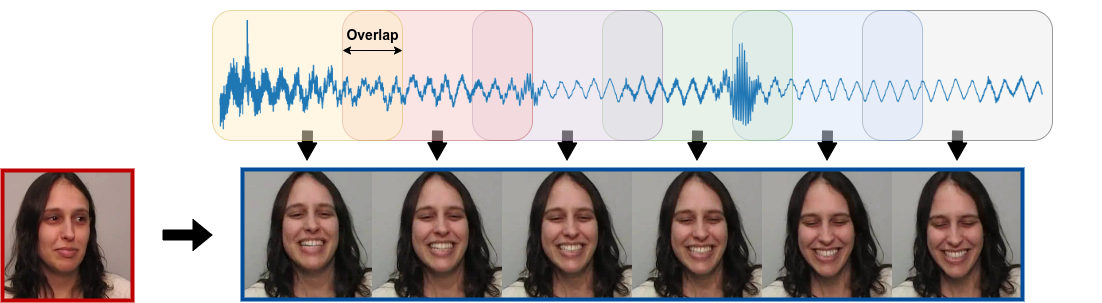}
\caption{The proposed end-to-end laughter generation model synthesizes a video of a laughing face using a still image of the speaker and an accompanying laughter segment. 
}
\end{figure}

\section{Introduction}
\label{sec:intro}
Facial animation is essential in many applications, such as virtual reality, movies, and human-computer interaction (HCI), by providing more immersive and engaging experiences. Current facial animation methods predominantly focus on speech-driven animation, resulting in the neglect of non-verbal expressions such as laughter, head nods, or blinks. This omission poses a substantial issue since these non-verbal cues often convey essential contextual information and play an important role in natural dialogue. Laughter is an interesting initial subject of study due to its ancient roots as a social signal~\cite{niemitz1990visuelle,ruch2001expressive, pentland2010honest}, acting as a powerful non-verbal communication medium that conveys emotions, intentions, and social relationships~\cite{provine2001laughter, glenn2003laughter}. However, laughter, unlike speech, lacks a direct correlation with lip movement. When combined with the scarcity of training data, this makes the development of a model for realistic laughter sequence generation quite challenging.

Until recently, speech-driven animation methods mainly relied on Generative Adversarial Networks (GANs)~\cite{goodfellow2020generative}. Early approaches were limited in terms of head rotations~\cite{vougioukas2018end} or could only modify lip movements~\cite{prajwalLipSyncExpert2020}. Recent advances have led to methods capable of generating realistic facial animations, with~\cite{liangExpressiveTalkingHead2022, zhou2020makelttalk, zhouPoseControllableTalkingFace2021a} or without~\cite{anim2} the use of intermediate representations such as key points, landmarks, or driving videos. Some of these methods even incorporate emotion control into the generation process~\cite{gururaniSPACExSpeechdrivenPortrait2022, jiEAMMOneShotEmotional2022}. The emergence of diffusion-based generation techniques has further spurred progress in the field, as researchers leverage the improved performance of these new models~\cite{shenDiffTalkCraftingDiffusion2023, stypulkowskiDiffusedHeadsDiffusion2023}. 
Current methods employ frame-based generators, exploiting the strong correlation between speech and lip movement. However, these models struggle with laughter generation due to several issues. Firstly, laughter lacks the robust audio-visual correlation seen in speech~\cite{chen2021audio, kadandale2022vocalist}, making the generation of authentic audio-driven laughter sequences considerably more difficult. Secondly, laughter's complexity and variability, involving various muscles and facial movements, poses a substantial challenge for existing frame-based generators. These models, which are designed for speech, primarily focus on the mouth and lips and struggle to capture the subtleties and variations in laughter, resulting in unnatural or inaccurate visual renditions. Finally, the spontaneity and context-dependency of laughter make it difficult to predict the timing and intensity of the speaker's facial movements accurately. These challenges emphasize the need for innovative approaches specifically designed for laughter generation.

In this paper, we design a novel video diffusion model to generate videos of laughing faces based on raw audio input. Our model leverages recent developments in video diffusion~\cite{ho2022video, singer2022make} to accurately capture the complex laughter dynamics, leading to realistic and synchronized laughing animations. To the best of our knowledge, our method is the first to generate natural laughter videos. In addition, we address the issue of limited publicly-available audio-visual laughter corpora by proposing an ensemble of existing datasets for training and evaluation purposes. To assess the quality of our results, we employ a series of metrics from existing video generation works and design a novel metric specifically tailored for laughter generation. We perform a thorough evaluation of our proposed method's performance and conduct an ablation study on our model to systematically assess the impact of each individual component within this system. We find that our approach outperforms previous state-of-the-art speech-driven facial animation models, including other diffusion-based methods, whether pre-trained on speech or re-trained on laughter. Furthermore, our method produces videos that are significantly better aligned with the input laughter audio.


\section{Related Work}

\textbf{Speech-Driven Facial Animation.} Early facial animation research~\cite{yehia1998quantitative} established a strong relationship between speech features and facial motion, initially leveraged through hidden Markov models (HMMs)~\cite{xie2007coupled, yamamoto1998lip}. Quality enhancements were realized with the rise of deep learning methods~\cite{karras2017audio, fan2015photo, suwajanakorn2017synthesizing}, notably through the introduction of generative adversarial networks (GANs)~\cite{goodfellow2020generative}. As a result, an initial wave of research focused on achieving lip synchronization~\cite{suwajanakorn2017synthesizing, lip3} with Prajwal \etal \cite{prajwalLipSyncExpert2020} attaining near-perfect synchronization for in-the-wild videos. Subsequent work incorporated natural facial expressions such as blinks and eyebrow movements but lacked head rotations~\cite{anim1, anim2}. Some methods addressed this by using intermediate representations like landmarks~\cite{zhou2020makelttalk, gururaniSPACExSpeechdrivenPortrait2022}, keypoints~\cite{jiEAMMOneShotEmotional2022} or a driving video~\cite{liangExpressiveTalkingHead2022, zhouPoseControllableTalkingFace2021a}. On the other hand, recent diffusion-based approaches~\cite{stypulkowskiDiffusedHeadsDiffusion2023} have demonstrated state-of-the-art performance, showcasing their ability to generate plausible head motion and diverse facial expressions by using only speech as a conditioning input. Other works have also focused on adding control over the emotion displayed in the generation~\cite{gururaniSPACExSpeechdrivenPortrait2022, liangExpressiveTalkingHead2022, chen2021audio}. Despite these remarkable advances, the generation of non-verbal aspects of human communication remains unexplored. 

\textbf{Diffusion Models.} 
Introduced in \cite{sohl2015deep, song2019generative, ho2020denoising}, diffusion models have shown strong generative capabilities in point cloud generation~\cite{luo2021diffusion}, music synthesis~\cite{schneider2023mo, huang2023noise2music} and video generation~\cite{singer2022make, ho2022video, ho2022imagen}. Compared to GANs, diffusion models provide a more stable and robust training process, as well as improved mode coverage which makes the model less likely to overfit~\cite{xiao2021tackling}. Score-based diffusion models, presented in \cite{song2020score} and improved in \cite{karras2022elucidating, dhariwal2021diffusion}, extend the original diffusion models by generalizing the noise distribution through the use of stochastic differential equations (SDEs). They can effectively capture complex data distributions and generate high-quality samples, while still maintaining the advantages of the original diffusion models in terms of denoising and sampling efficiency. These advancements allow for a broader range of applications and adaptability to different domains. More recently, \cite{rombach2021highresolution} propose Latent Diffusion Models, managing to produce high-resolution images by transferring the
training and inference processes to a compressed
lower-dimension latent space for more efficient computing. However, we found that this approach failed to yield successful results in our case, likely due to the limited amount of data available.


\textbf{Laughter in Human Communication.} Laughter generation has been explored across various modalities, including audio, text, and animation. In the audio domain, studies have primarily focused on synthesizing laughter sounds~\cite{urbain2013automatic, lasarcyk2007imitating, tits2020laughter} to extend the capabilities of text-to-speech systems. In the text domain, researchers have investigated methods for generating and recognizing laughter in textual conversations, such as identifying and generating laughter events in dialogues~\cite{bertero2016long}. For animation, Ding \etal.~\cite{ding2014laughter} developed a real-time laughter animation generator that takes input pseudo-phonemes of laughter and their duration times, synthesizes facial and body motions by learning the relationship between input signals and human motions, and employs a combination of contextual Gaussian Models and motion capture data. More recently, projects like ILHAIRE~\cite{dupont2016laughter} have showcased the importance of laughter synthesis and recognition in human-avatar interactions. Incorporating laughter in facial animation is crucial for developing more realistic and engaging virtual characters, ultimately enhancing the overall user experience.

\section{Methodology} 

\begin{figure}[t]
\centering
\includegraphics[width=\linewidth]{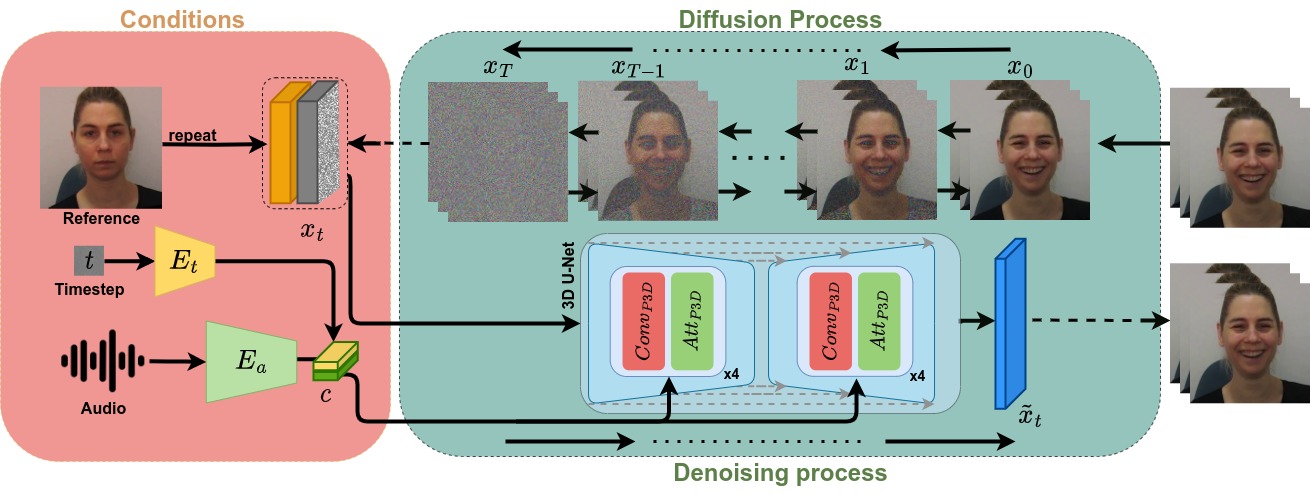}
\caption{Overview of our proposed pipeline for laughter generation. The model takes as input the noisy video concatenated with the reference frame and outputs the denoised version of the images conditioned on the laughter audio and the timestep of the diffusion process.}
\label{fig:arch}
\end{figure}

\subsection{Diffusion models}

Diffusion models~\cite{ho2020denoising, sohl2015deep, song2019generative} are a class of generative models that synthesize samples by progressively removing random noise. The input to a conditional diffusion model consists of a conditioning signal $c$, a random time step $t$, and a sample $x_t$ obtained by corrupting the original data $x$ by adding i.i.d. Gaussian noise of standard deviation $\sigma$. 

We adopt the approach of Karras \etal~\cite{karras2022elucidating} that further explores the design choices of this type of model, both theoretically and empirically, and presents a sampling process that uses Heun's method as the ODE solver, reducing the number of neural function evaluations needed while maintaining the FID score~\cite{croitoru2023diffusion}. This process is characterized by a noise schedule with a standard deviation $\sigma_t$ at time $t$. The time range $t$ is uniformly sampled during training, with the diffusion progressing in the direction of increasing time. The Gaussian diffusion dynamics can be fully described by a single noise vector $\pmb{n}\sim\mathcal{N}(\pmb{0},\sigma^2\pmb{I})$ with noise levels $\sigma_0=\sigma_{max}>\sigma_1>\dots>\sigma_T=0$, as $x_t$ can be expressed as a function of the original sample and the noise vector $\pmb{n}$, i.e., $x_t = x + \pmb{n}$. The model $D_\theta$ is trained to determine the original image given this input. The diffusion loss minimizes the expected $L_2$ denoising error for samples drawn from the training data separately for every $\sigma$, i.e.:

\begin{equation}
\mathcal{L}(\theta) = \mathbb{E}_{x, c, t, \sigma}[w_t\lVert D_\theta(x_t;c, \sigma_t) - x \rVert^2_2],
\end{equation}
where $w_t$ is a fixed weight function of choice. Inference is performed by taking random noise at time $t_{max}$ and denoising it using the noise predictions provided by the model. 



\subsection{Architecture}
\begin{figure}[ht]
\centering
\includegraphics[width=\linewidth]{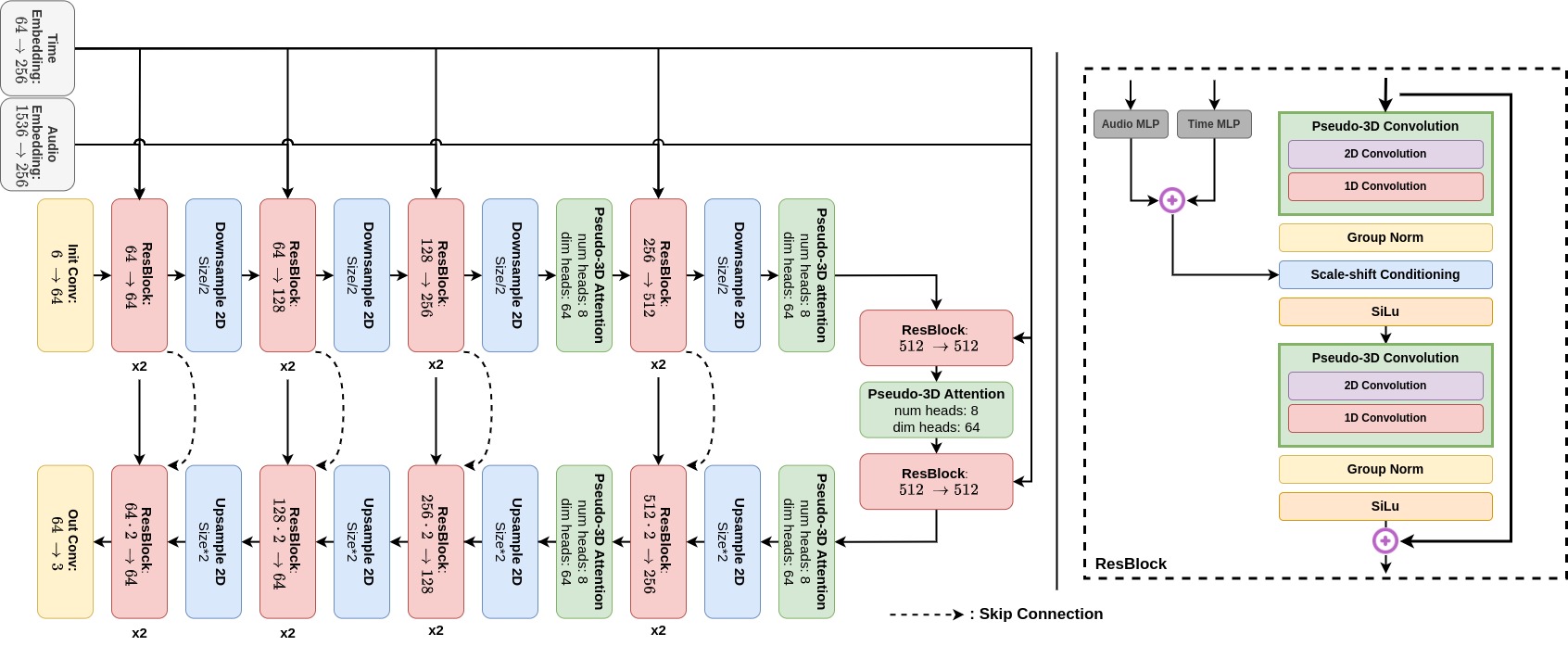}
\caption{Details of the U-Net layers. Left: Overview of the U-Net architecture. Right: Detailed view of the ResBlock.}
\label{fig:arch_detail}
\end{figure}

Our architecture primarily builds upon the work of Ho \etal~\cite{ho2022video}, which employs a factorized space-time U-Net architecture, extending the standard 2D U-Net used in image diffusion models. The model, illustrated in Fig.~\ref{fig:arch}, comprises four down-sampling and up-sampling blocks connected by residual connections. The input of our model is a video sample $x_t \in \mathbb{R}^{B \times C \times F \times H \times W}$, where $B,C,F,H,W$ are the batch, channels, frames, height, and width dimensions respectively. The condition signal $c$, in our case, consists of a single frame $x'\in \mathbb{R}^{B \times C \times H \times W}$ concatenated channel-wise with $x_t$ by repeating it in the temporal dimension, and an audio sequence $a$ which is passed through an audio encoder ($E_a$). Additionally, we pass the timestep information of the diffusion process $t$ processed by a two-layer MLP ($E_t$). The U-Net contains a composition of convolutional and self-attention layers followed by a down-sampling or up-sampling layer. In our proposed method, we apply Pseudo-3D Convolutional and Attention Layers~\cite{singer2022make} to balance computational efficiency, and information sharing in the network. For each layer, instead of using the full 3D convolution, we use a 2D convolution applied to the spatial dimensions ($\mathbb{R}^{B \times C \times F \times H \times W} \rightarrow \mathbb{R}^{(B \times F) \times C \times H \times W}$), followed by a  1D convolution applied to the temporal dimension by merging the other dimensions ($\mathbb{R}^{B \times C \times F \times H \times W} \rightarrow \mathbb{R}^{(B \times H \times W) \times C \times F}$). We apply a similar strategy for the attention layers. The details of the network can be found in Fig.~\ref{fig:arch_detail}.


The audio signal is processed using an audio encoder from \cite{beats} pre-trained on AudioSet~\cite{audioset}. We split the corresponding audio sequence into chunks of equal length based on the number of frames in the video, resulting in a new audio sequence $a' = \{a'_0, ... ,a'_F\}$. Then, the audio and timestep conditioning is performed in each of the ResBlocks of the U-Net at the first convolutional layer. This is done by modulating the input $x_t$ through a scale-shift operation after a group normalization layer ($GN$):
\begin{equation}
h_{s+1} = GN(h_s) * (k + 1) + s
\end{equation}
where $h_s$ and $h_{s+1}$ are consecutive hidden states of the U-Net, and $k$ and $s$ are the scale and shift, respectively. To obtain $k$ and $s$, we sum the audio sequence $a_t$ and the encoded timestep information $t$ and pass it through a linear layer. We then split the result to obtain the scale and shift.


\subsection{Training} \label{sec:training}

One challenge in generating laughter is the lack of large publicly available datasets. To mitigate the risk of overfitting, a common issue in diffusion models trained on smaller datasets, we adopt a series of techniques described below that we further discuss in section \ref{sec:abla}:

\textbf{Augmentation regularization.} We use a technique originally developed for training GANs with limited data~\cite{karras2020training} and later successfully applied to diffusion models~\cite{ karras2022elucidating}. The pipeline incorporates several geometric transformations which are applied to training images prior to adding noise. To prevent these augmentations from leaking into the generated images, we supply the augmentation parameters as conditioning input to $D_\theta$. During inference, we set these parameters to zero, ensuring only non-augmented images are generated.

\textbf{Classifier-free guidance (CFG).} This technique, proposed by Ho and Salimans~\cite{ho2022classifier}, enhances the alignment between generated samples and conditional input. At inference, the noise vector is computed both with and without the conditional input, and the final noise vector is determined by $wD_\theta(x_t;c, \sigma_t) + (1 - w)D_\theta(x_t;\cdot, \sigma_t)$, where $w > 1$. We found that $w=1$ worked best in our case. During training, the conditional input is masked with a probability of 0.1, enabling the same model to handle both conditional and unconditional generation.


\textbf{Longer sequence generation.} Due to computational constraints, we train our model on sequences of 16 consecutive frames randomly sampled from the videos, rather than training on the full videos. However, during inference, we can generate arbitrary lengths by autoregressively sampling our model. Initially, we generate a video $x_a \sim p_\theta(x)$ and use the last frame of $x_a$ as a condition for $x_b \sim p_\theta(x_b|x_a)$. 

We train all models using the Lion optimizer~\cite{chen2023symbolic} with $\beta_1=0.95$ and $\beta_2=0.98$ and a learning rate of $6\times10^{-5}$. During initial experimentation, we found that Lion converged noticeably faster than commonly used optimizers such as Adam~\cite{kingma2014adam} or AdamW~\cite{LoshchilovH19}, while consistently achieving equivalent or superior final performance. We linearly warm up the learning rate for the first 20 epochs and subsequently apply a cosine decay schedule until the end of training. We train our models for 200 epochs with a total minibatch of 32 samples. 

\section{Experiments}

\subsection{Datasets}\label{sec:dataset}

We identified four datasets suitable for laughter generation, namely MAHNOB~\cite{petridis2013mahnob}, AVLaughterCycle~\cite{urbain2010avlaughtercycle}, AVIC~\cite{schuller2009being} and SAL~\cite{douglas2008sensitive}. As these datasets do not solely contain laughter, we focus on the videos that feature it. We divide the data into training, validation, and test sets following an 80 -- 10 -- 10\,\% ratio, ensuring there is no overlap between the speakers in each set. The exact split of data can be found in the supplementary material. For all datasets, we use an audio sampling rate of 16\,kHz and a video frame rate of 25. During preprocessing, we align all faces to a canonical face and normalize images to the [-1,1] range. Details for each dataset are presented in Table~\ref{tab:dataset}.


\begin{table}[ht]
\centering

\resizebox{\linewidth}{!}{
\begin{tabular}{lrrrr}
\toprule
Dataset         & \# Speakers & \# Videos & Avg. length (sec.) & Total length (hours)  \\ \midrule
AVLaughterCycle~\cite{urbain2010avlaughtercycle} & 8          & 421      & 3.80 $\pm$  6.42                & 0.44                      \\
Mahnob~\cite{petridis2013mahnob}        & 22         & 554      & 1.56  $\pm$ 2.21                & 0.24                      \\
AVIC~\cite{schuller2009being}          & 21         & 312      & 0.36      $\pm$ 0.30                & 0.03                      \\
SAL~\cite{douglas2008sensitive}         & 28         & 98       & 1.46         $\pm$ 0.77                & 0.04 \\
\bottomrule
\end{tabular}
}
\vspace{0.2cm}
\caption{Overview of the datasets used in the study.}
\label{tab:dataset}
\end{table}


\subsection{Evaluation Metrics} \label{sec:metrics}
We employ widely-used reconstruction metrics, such as peak Frechet Inception Distance (FID) and structural similarity (SSIM) index, to assess the quality of generated images. Furthermore, we employ Frechet Video Distance (FVD) to evaluate visual quality, temporal coherence, and sample diversity. To assess the authenticity of the generated laughing faces, we train a Laughter Classifier (LC) to differentiate between speech and laughter videos. This model, based on a pre-trained MViTv2~\cite{LiW0MXMF22} for video classification on Kinetics~\cite{kinetics}, is fine-tuned with laughter and speech data from MAHNOB~\cite{petridis2013mahnob}. More details are available in the supplementary material. This novel metric allows us to highlight the limitations of pre-trained speech-driven animation methods, while simultaneously demonstrating our model's capability to generate realistic laughter sequences. The Laughter Classifier achieves an accuracy and F1 score of 100\,\% on the test set. We then apply this trained model to categorize the generated videos, assessing whether they are accurately classified as laughter. 
When measuring the Laughter Classifier's accuracy, we prevent any bias caused by the initial frame by ensuring that the reference frame is a neutral face, which may not always be the case when sampling a random video. This is crucial as a smiling face can easily resemble laughter, introducing bias into our evaluation.



\section{Results}
To the best of our knowledge, this is the first work on audio-driven laughter generation, so we compare against three speech-driven animation methods that we re-trained for laughter: Diffused Heads~\cite{stypulkowskiDiffusedHeadsDiffusion2023}, SDA~\cite{vougioukas2018end}, and EAMM~\cite{jiEAMMOneShotEmotional2022}. We also compare with pre-trained models such as MakeItTalk~\cite{zhou2020makelttalk} and PC-AVS~\cite{zhouPoseControllableTalkingFace2021a}. Furthermore, we perform ablation studies on various design choices within our framework and discuss their importance. In terms of video generation, we evaluate the models at a resolution of $128\times128$. These models are conditioned on a single frame and generate the following 16, adhering to the FVD model's expectation of a 16-frame video. However, for human evaluations, we opted to generate 2-second videos.

\subsection{Comparison with Other Works} \label{sec:comparison}

\begin{table}[t]
\centering
\begin{tabular}{lccccc}
\toprule
Model       & FVD $\downarrow$    & FID $\downarrow$ & SSIM $\uparrow$   & LC $\uparrow$ (\%)  & MOS $\uparrow$  \\ \midrule
\rowcolor{lightgray!30}
\multicolumn{6}{c}{\textit{Pre-trained}} \vspace{.5em}\\
Diffused Heads~\cite{stypulkowskiDiffusedHeadsDiffusion2023} &  {149.51}   & {49.36}   & 0.236  &     80.70            &       \multicolumn{1}{c}{-}               \\
SDA~\cite{vougioukas2018end}        & 594.32   &  111.89 & 0.053 &   13.85              &      \multicolumn{1}{c}{-}             \\
 EAMM~\cite{jiEAMMOneShotEmotional2022}            &  391.62  &  71.71  & 0.094  &    16.67             &    \multicolumn{1}{c}{-}               \\ 
 PC-AVS~\cite{zhouPoseControllableTalkingFace2021a}            &  1164.49  & 175.99  & 0.004 &      53.91           & 
  \multicolumn{1}{c}{-}           \\ 
 MakeItTalk~\cite{zhou2020makelttalk}            & 196.89   &  49.08 & 0.262 &          72.50          &   1.94$\pm$1.12            \\ \midrule
\rowcolor{lightgray!30}
\multicolumn{6}{c}{\textit{Re-trained}} \vspace{.5em}\\
Diffused Heads~\cite{stypulkowskiDiffusedHeadsDiffusion2023} &  152.30   & 67.46  & 0.232 &     {94.09}               &  2.45$\pm$1.22          \\
  SDA~\cite{vougioukas2018end}          & 696.33   &  124.52 & 0.040 &   85.13              &       -               \\
 EAMM~\cite{jiEAMMOneShotEmotional2022}            &  324.97  & 74.18  & 0.095 &         20.67        &        1.87$\pm$1.05        \\
 \midrule
 Laughing Matters (Ours)             &  \textbf{111.95}  &  \textbf{45.69}  & \textbf{0.371} &   \textbf{96.52}              &    \textbf{3.39$\pm$1.09}             \\
 \midrule
Ground truth &  \multicolumn{1}{c}{-}   & \multicolumn{1}{c}{-}  & \multicolumn{1}{c}{-} &   100.00  &   3.49$\pm$1.23 \\
\bottomrule
\end{tabular}
\vspace{2mm}
\caption{Comparative performance of the proposed methods against pre-trained and re-trained models. The best result is highlighted in \textbf{bold}.}
\label{tab:res_retrain}
\end{table}

As shown in Table~\ref{tab:res_retrain}, models that are pre-trained on speech struggle to generate satisfactory results, especially in terms of the Laughter Classifier metric. This highlights the need for re-training the models on laughter. Consequently, we also compare our method against re-trained models, adhering to the recommended parameters from their respective papers. Despite the improvements achieved by re-training, our approach consistently outperforms other methods in terms of visual quality and laughter accuracy. We primarily attribute our model's performance to our 3D architecture, which, unlike other frame-based methods, enables longer audio context. This is essential since laughter exhibits a lower correlation between acoustic and visual cues compared to speech. Other significant improvements stem from the choice of an audio encoder specifically tailored for laughter and the training improvements employed to compensate for limited training data, as discussed in Section \ref{sec:abla}.

We further validate our model's superior performance through a Mean Opinion Score (MOS) test. Participants are shown an average of 23 randomly selected videos, featuring a blend of ground truth, our model, Diffused Heads~\cite{stypulkowskiDiffusedHeadsDiffusion2023}, MakeItTalk~\cite{zhou2020makelttalk}, and EAMM~\cite{jiEAMMOneShotEmotional2022}. Participants watch the videos sequentially and rate them on a scale of 1 to 5, where 1 indicates the video appears clearly artificial, and 5 suggests it is highly realistic and indistinguishable from genuine laughter. We collect a total of 72 responses that we detail in the supplementary material. Even though the performance difference is minimal in the laughter metric, it is significant in terms of user preference, where temporal smoothness and natural expressions are crucial factors. It is worth highlighting that even ground truth videos score relatively low, which is likely due to the difficulties in assessing whether a laughter video is realistic and well synchronized with its corresponding audio. This discriminative task is indeed challenging, even for humans, as evidenced by the videos shown on the project page.

Additionally, we compare two variants of our approach as two different models in the user study: with and without head rotations. This is achieved by taking the original video and eliminating the head rotation using the model from \cite{DrobyshevCKILZ22}. We evaluate both models on videos taken from the MAHNOB~\cite{petridis2013mahnob} test set. The results, shown in Table~\ref{tab:mos_rot}, show that removing head rotation severely deteriorates performance, highlighting the importance of correctly modelling the speaker's head movements when generating laughter videos.

\begin{table}[ht]
\centering
\begin{tabular}{lr}
\toprule
Model      &  MOS $\uparrow$ \\ \midrule
Laughing Matters w/ rotations &  3.08 $\pm$ 1.12   \\
Laughing Matters w/o rotations &    2.08  $\pm$ 1.07 \\
\bottomrule
\end{tabular}
\vspace{2mm}
\caption{Mean Opinion Score of our model with and without the head rotations.}
\label{tab:mos_rot}
\end{table}

\subsection{Ablation study}\label{sec:abla}



\textbf{Audio Encoder.} Choosing the right audio encoder is essential for achieving optimal results. While a pre-trained model on a large speech dataset is a common choice for speech animation, they prove inadequate for our specific use case. As indicated by Table~\ref{tab:abl_audio_enc}, speech encoders such as SDA~\cite{vougioukas2018end} and WavLM~\cite{wavlm} yield unsatisfactory results, producing outputs closer to speech rather than laughter, as observed in the Laughter Classifier metric. Training from scratch, for instance with mel-spectrograms, provides some improvement as it allows the model to learn directly from the laughter data. However, due to the limited availability of training data, it is highly beneficial to identify a pre-trained model suitable for our task. To this end, we apply BEATs~\cite{beats}, a state-of-the-art self-supervised audio encoder. Being trained on AudioSet~\cite{audioset}, which contains 15.8 hours of laughter data, it achieves superior performance across all metrics. 

\begin{table}[ht]
\centering
\begin{tabular}{lrrrr}
\toprule
Audio Encoder   & FVD $\downarrow$   & FID $\downarrow$  & SSIM $\uparrow$   & LC $\uparrow$ (\%)   \\ \midrule
SDA~\cite{vougioukas2018end} &     169.48      &   55.07    &      {0.318}        &     68.21                            \\
WavLM~\cite{wavlm}          &  136.76        &   {46.01}    &       0.312          &        54.21                     \\
Mel-spectrograms          &    {124.81}      &   47.74     &      0.320           &       {83.52}       \\
BEATs~\cite{beats}           &  \textbf{111.95}         &  \textbf{45.69}     &      \textbf{0.371}          &     \textbf{96.52}                         \\
\bottomrule
\end{tabular}
\vspace{2mm}
\caption{Ablation study on the audio encoder. }
\label{tab:abl_audio_enc}
\end{table}

\textbf{Training improvements.} To mitigate overfitting, a common issue with diffusion models trained on smaller datasets, we implement two techniques detailed in Section \ref{sec:training}: Augmentation regularization and Classifier-free guidance (CFG). Table~\ref{tab:abl_train} illustrates the impact of both components, demonstrating consistent performance improvements when they are used.

\begin{table}[ht]
\centering
\begin{tabular}{lrrrr}
\toprule
Training configuration & FVD $\downarrow$ & FID $\downarrow$  & SSIM $\uparrow$   & LC $\uparrow$ (\%)   \\ \midrule
Baseline &  \textbf{111.95}         &  \textbf{45.69}     &      \textbf{0.371}          &     \textbf{96.52}                     \\
w/o Augmentation regularization         &   195.03       &    60.60   &   0.308              &        83.93              \\
w/o Classifier-free guidance          &   126.89        &   46.91    &              0.302   &      75.09                  \\
\bottomrule
\end{tabular}
\vspace{2mm}
\caption{Ablation study on the training improvements.}
\label{tab:abl_train}
\end{table}

\subsection{Qualitative evaluation}
\begin{figure}[ht]
\centering
\begin{minipage}{.49\textwidth}
  \centering
  \includegraphics[width=.72\linewidth]{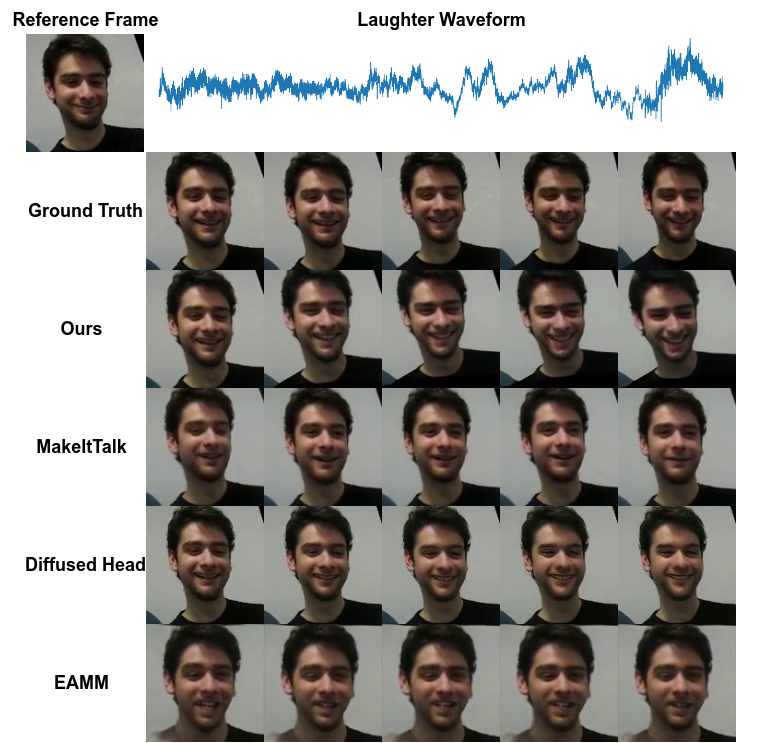}
  \caption{Qualitative evaluation results. The reference frame and the laughter waveform can be seen on the top.}
  \label{fig:qualitative}
\end{minipage}
\hspace{0.02\linewidth}
\begin{minipage}{.42\textwidth}
  \centering
  \includegraphics[width=.95\linewidth]{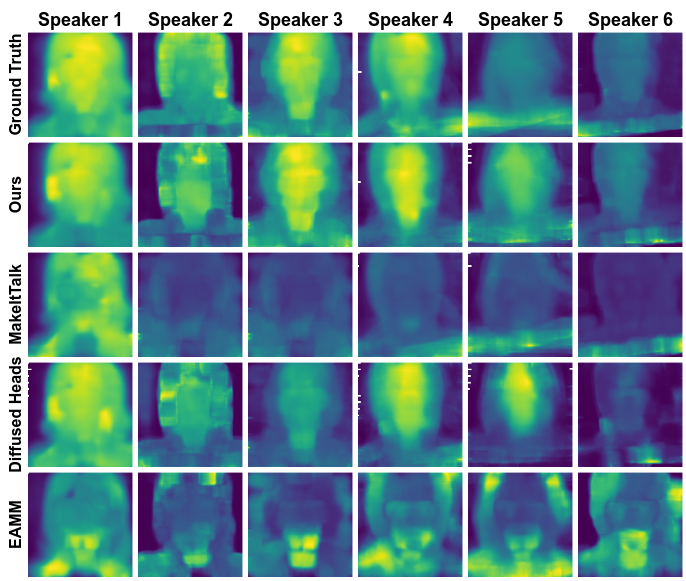}
  \caption{Average magnitude of optical flow for different speakers across all videos.}
  \label{fig:optical_flow}
\end{minipage}%
\end{figure}

Our method effectively generates realistic videos from previously unseen faces and audio clips taken from the test set. Fig.~\ref{fig:qualitative} illustrates the laughter sequence generated by our model and competing approaches. Upon visual examination, it is apparent that EAMM~\cite{jiEAMMOneShotEmotional2022} struggles to preserve identity, whereas MakeItTalk~\cite{zhou2020makelttalk} only animates the lips. While Diffused Heads~\cite{stypulkowskiDiffusedHeadsDiffusion2023} can consistently produce high-quality visuals, the synchronization with the audio input often falls short. Conversely, our model succeeds in creating a laughter sequence with correlated head movement. For a deeper understanding of our results, we invite readers to review the generated videos available in the supplementary material, where our model generates various laughter types and ad-
justs to out-of-distribution speakers. 

Moreover, our aim is to demonstrate that our model can replicate the movement patterns seen in real laughter videos. Fig.~\ref{fig:optical_flow} presents a comparison of the average magnitude of optical flow for various speakers, showing the regions of the frames that exhibit the most movement. The heatmaps from our generated videos closely align with the ground truth across all speakers, validating our model's ability to create laughter sequences with natural movement. In contrast, MakeItTalk~\cite{zhou2020makelttalk} and EAMM~\cite{jiEAMMOneShotEmotional2022} yield results that significantly deviate from the ground truth. Notably, while Diffused Heads~\cite{stypulkowskiDiffusedHeadsDiffusion2023} generates somewhat accurate movements, it falls short in matching the ground truth for Speakers 3, 5, and 6.

\section{Conclusion}

In this work, we introduce Laughing Matters, an end-to-end model that synthesizes realistic laughing faces from a still image and an audio clip. Our approach outperforms existing methods in generating convincing laughter animations, as demonstrated through quantitative and qualitative evaluations. 
We conduct a set of ablation studies to examine the impact of the audio encoder and training improvements. Our findings reveal that using a laughter-specific audio encoder, applying augmentation regularization techniques, and leveraging classifier-free guidance significantly enhance the model's performance. 
Looking forward, it would be promising to extend our model to cover other non-verbal cues, with the aim of creating a comprehensive facial animation model that can animate all verbal and non-verbal cues present in natural speech.
\bibliographystyle{unsrt}  
\bibliography{refs}

\newpage
\appendix

\section{Datasets}

\subsection*{Training split}

The models are trained and evaluated on MAHNOB, AVLaughterCycle, AVIC and SAL. The speakers used for training, validation and testing are shown in Table \ref{tab:dataset_split}. 

\begin{table}[ht]
\centering

\resizebox{\linewidth}{!}{
\begin{tabular}{lrrr}
\toprule
Dataset         &  Training & Validation & Testing  \\ \midrule
AVLaughterCycle &  11,13,18,5,6,7    & 16 & 14                \\
Mahnob        &  1,2,4,6,7,8,9,11,13,15,17,19,20,22,23,24,25  & 5,16 &  3,14,21        \\
AVIC         & 4,5,6,8,9,10,12,13,26,27,30,31,32,33,34,35  & 16,36 & 15,28,29             \\
SAL & \begin{tabular}{@{}r@{}}Alex, Donn, Gary, Liam, Mlind,\\ Nicol, Ruth, Shar, Ed, Ian, Rod\end{tabular} & Mart, GHill & Nol, Alis \\
\bottomrule
\end{tabular}
}
\vspace{0.1mm}
\caption{Speaker IDs for training, validation and test sets.}
\label{tab:dataset_split}
\end{table}

\section{Laughter Classifier}

We introduce the Laughter Classifier as part of our evaluation methodology. This not only highlights the limitations of pre-trained speech-driven animation methods, but also demonstrates the capabilities of our model in generating realistic laughter sequences. The model processes video inputs and produces a single logit output to classify whether the individual in the video is speaking or laughing.

\subsection{Architecture}

\begin{figure}[ht]
\centering
\includegraphics[width=0.8\linewidth]{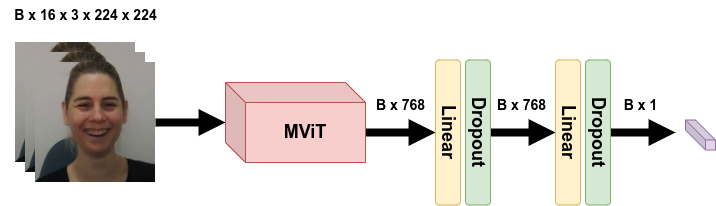}
\caption{The architecture used for the Laughter Classifier. The batch size is denoted as B.}
\label{fig:laughter_classifier_arch}
\end{figure}

The architecture of the system is presented in Fig. \ref{fig:laughter_classifier_arch}. We employ a Multiscale Vision Transformers (MViT) \cite{LiW0MXMF22} backbone with two linear layers and a dropout layer with a dropout probability set at 0.2. The \href{https://pytorch.org/vision/stable/models/generated/torchvision.models.video.mvit_v2_s.html#torchvision.models.video.mvit_v2_s}{MViT} model, pre-trained on the Kinetics 400 dataset, reaches a top-5 accuracy of 94.665\,\%.

\subsection{Training}

Our model was trained using the MAHNOB training set for laughter combined with additional speech videos for each speaker, given that MAHNOB also provides speech data. During the training process, either laughter or a speech video is fed into the model with equal probability. We train with the AdamW optimizer, configured with a learning rate of $1\times10^{-4}$, $\beta_1=0.9$, and $\beta_2=0.999$. Binary Cross Entropy is employed as the loss function.

\section{Study on CFG scale}

\begin{figure}[ht]
\centering
\includegraphics[width=0.8\linewidth]{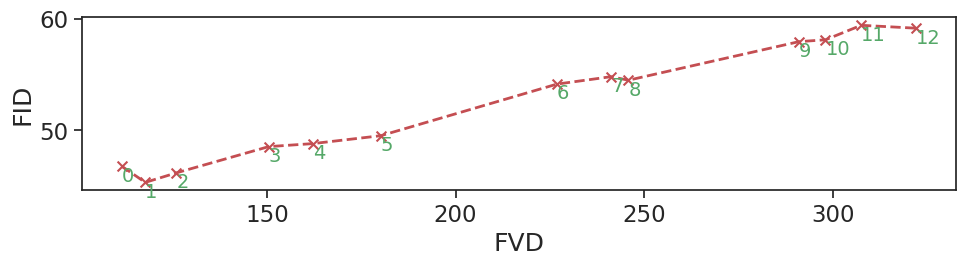}
\caption{CFG analysis by comparing FID and FVD for different scales}
\label{fig:cfg}
\end{figure}

We investigated the effects of the CFG scale on the output of our diffusion model. The CFG scale is a parameter that controls how closely the generated image adheres to the user's condition. We find that a higher CFG scale value resulted in images that were more faithful to the condition, but also had more artefacts. A lower CFG scale value resulted in images with less noise, but they were also less faithful to the condition. We show in Fig.~\ref{fig:cfg} a comparison between FID and FVD for scales between 0 and 12 and found CFG=1 to be the optimal tradeoff in our case.

\section{User study}

\begin{figure}[ht]
\centering
\includegraphics[width=0.60\linewidth]{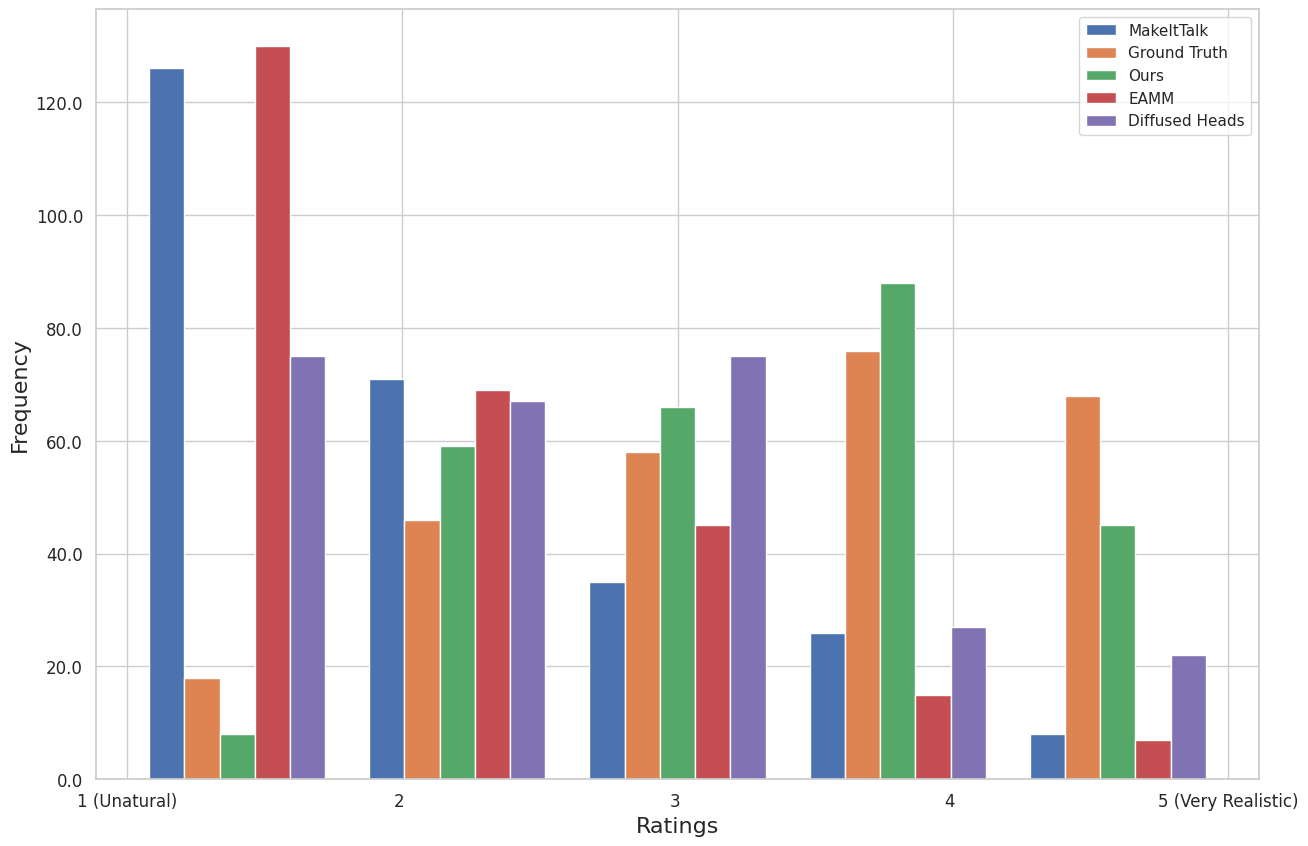}
\caption{Histogram of the user study rating for our model compared to MakeItTalk~\cite{zhou2020makelttalk}, EAMM~\cite{jiEAMMOneShotEmotional2022}, Diffused Heads~\cite{stypulkowskiDiffusedHeadsDiffusion2023} and Ground Truth.}
\label{fig:user}
\end{figure}

\begin{figure}[ht]
\centering
\includegraphics[width=0.55\linewidth]{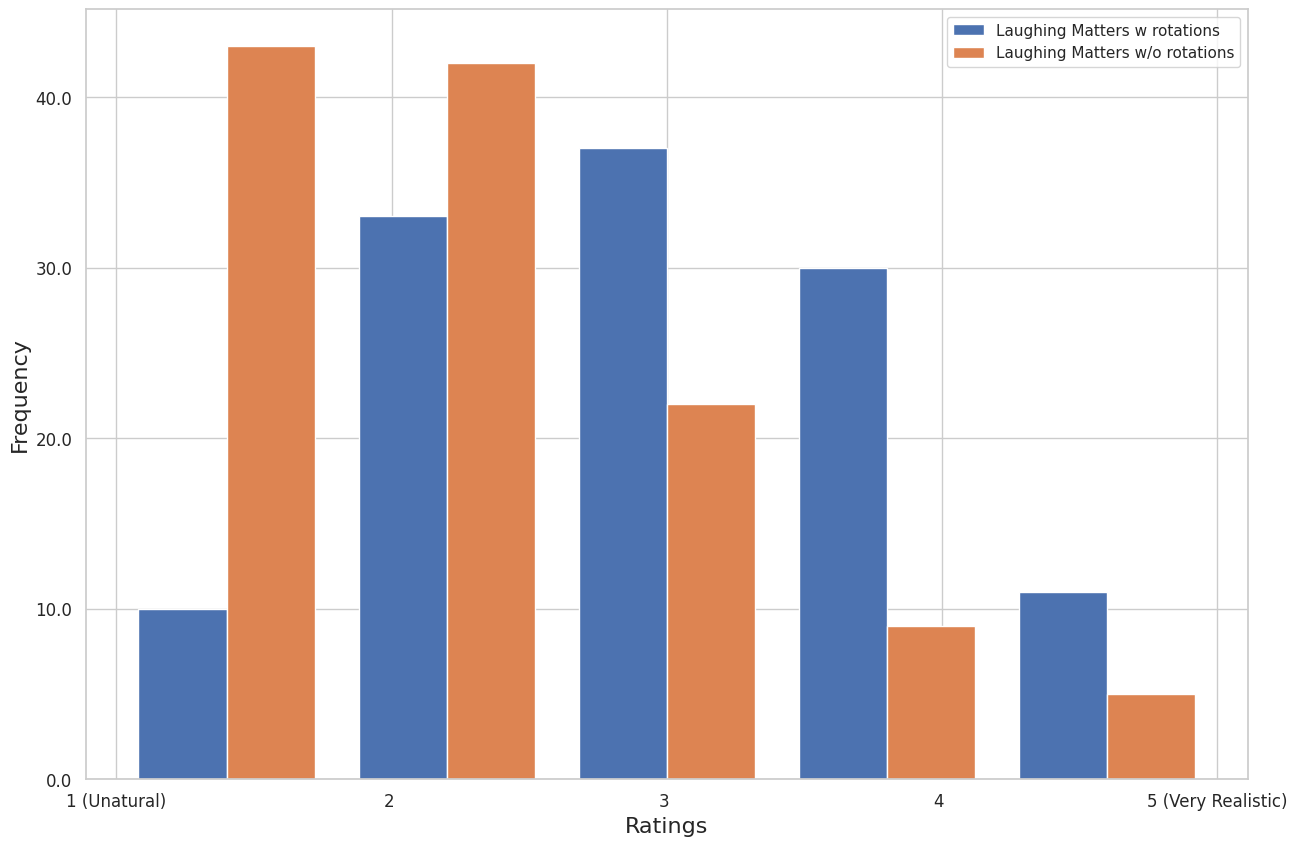}
\caption{Histogram of the user study rating for our model with and without head rotations.}
\label{fig:rot_user}
\end{figure}

Since quality metrics do not always align with human perception, we conduct a Mean Opinion Score evaluation. In this study, participants rate videos from different models on a scale of 1 to 5, with 1 indicating that the video appears clearly artificial, and 5 signifying that it is highly realistic and indistinguishable from genuine laughter.
Each participant views a minimum of 12 videos, with the option to extend their participation up to 60 videos. We have gathered 72 responses in total, averaging 23 videos per individual participant. The distribution of responses for each model is depicted in Fig. \ref{fig:user} and Fig. \ref{fig:rot_user}.

\section{Limitations}

\begin{figure}[ht]
\centering
\includegraphics[width=0.65\linewidth]{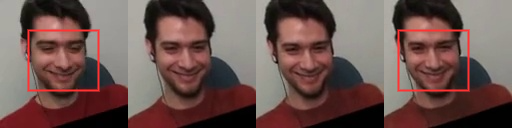}
\caption{Failure case of our network.}
\label{fig:limit}
\end{figure}

Throughout the evaluation process, we identified potential issues related to long-term generation, attributed to the autoregressive nature of the process and the limited data availability. Generating sequences longer than 2 seconds (or 50 frames) resulted in degraded quality. An example of such failure case for longer generation sequences can be observed in Fig. \ref{fig:limit}. In our specific use case, this is not a major concern, as laughter is typically brief. For instance, the average laughter duration in the MAHNOB dataset is 1.56 seconds, suggesting that our model can handle the majority of laughter episodes.  However, this highlights a valuable direction for future research. One potential solution could involve conditioning the model via an additional, unchanging \textit{identity frame}. This strategy could provide the model with enough information to maintain consistent quality throughout the autoregressive generation process.

\section{Videos}

For a more comprehensive understanding of our findings, we encourage readers to examine the side-by-side comparison video provided as part of the supplementary material. Within the same video, we also demonstrate a comparison between voiced and unvoiced laughter, highlighting our model's ability to generate both types. Voiced laughter is characterized by its harmonically rich, vowel-like sound, accompanied by measurable periodicity in vocal fold vibration. On the other hand, unvoiced laughter is produced through a noisy exhalation from either the nose or mouth, without involvement of the vocal folds. This distinction is crucial as research has shown that these two forms of laughter serve different functions in social interactions \cite{petridis2013mahnob}.

\end{document}